\documentclass[letterpaper]{article}

\usepackage{natbib,alifeconf}  

\usepackage{url}
\usepackage{booktabs}
\usepackage{hyperref}
\usepackage{cleveref}
\usepackage{stfloats}

\title{Hypernetworks That Evolve Themselves}

\author{
     Joachim Winther Pedersen$^{1}$,
    Erwan Plantec$^{1}$, 
    Eleni Nisioti$^{1}$, 
    Marcello Barylli$^{1}$,\\
    {\large Milton Montero$^{1}$,
    Kathrin Korte$^{1}$, 
    \and
    Sebastian Risi$^{1,2}$}\\
    \mbox{}\\
    $^1$IT University of Copenhagen, Denmark \\
    $^2$ Sakana AI\\
    jwin@itu.dk, sebr@itu.dk
} 

\begin{document}

\maketitle
\textbf{}
\begin{abstract}
How can neural networks evolve themselves without relying on external optimizers? We propose \emph{Self-Referential Graph HyperNetworks}, systems where the very machinery of variation and inheritance is embedded within the network. By uniting hypernetworks, stochastic parameter generation, and graph-based representations, Self-Referential GHNs mutate and evaluate themselves while adapting mutation rates as selectable traits. Through new reinforcement learning benchmarks with environmental shifts (CartPole-Switch, LunarLander-Switch), Self-Referential GHNs show swift, reliable adaptation and emergent population dynamics. In the locomotion benchmark Ant-v5, they evolve coherent gaits, showing promising fine-tuning capabilities by autonomously decreasing variation in the population to concentrate around promising solutions. Our findings support the idea that evolvability itself can emerge from neural self-reference. Self-Referential GHNs reflect a step toward synthetic systems that more closely mirror biological evolution, offering tools for autonomous, open-ended learning agents.
\end{abstract}


Data/Code available at: \url{https://github.com/Joachm/self-referential_GHNs}

\section{Introduction}

Neural networks have achieved remarkable success across diverse domains such as image recognition \citep{krizhevsky2012imagenet, he2016deep}, natural‑language understanding \citep{devlin2019bert, brown2020language}, and strategic decision‑making in games \citep{silver2016mastering}. This progress is powered largely by gradient‑based optimization procedures, including back‑propagation \citep{rumelhart1986learning}. Nevertheless, gradient methods exhibit persistent shortcomings—susceptibility to local minima, ill‑conditioned landscapes, and an inability to cope with non‑differentiable objectives or discrete architectural choices \citep{goodfellow2016deep, choromanska2015loss}.

These limitations have sparked a renewed wave of interest in evolutionary algorithms (EAs) as an alternative or complementary approach to optimizing neural networks \citep{rechenberg1984evolution, holland1984genetic, back1996evolutionary, risi:book25}. Modern large‑scale instantiations, including OpenAI Evolution Strategies \citep{salimans2017}, Deep Neuroevolution \citep{such2017deep}, and population‑based neural architecture search \citep{real2019regularized, stanley2019designing} demonstrate that, when supplied with massive parallel compute, EAs can match or exceed gradient learners on reinforcement‑learning and AutoML benchmarks. Their derivative‑free, population‑based nature also makes them well‑suited to distributed training regimes \citep{jaderberg2017population, salimans2017}.

A lesson from evolutionary biology is that the rate and structure of mutation matter: low variation accelerates exploitation but risks premature convergence, whereas high variation fosters exploration but risks already accrued fitness \citep{giraud2001costs}. Adaptive mutation rates, as heritable traits of individuals, might be especially important in open-ended systems, where the fitness landscape might undergo severe changes over time \citep{lehman2025evolution}. 

\begin{figure*}[htbp!]
    \centering
    \includegraphics[width=0.8\linewidth]{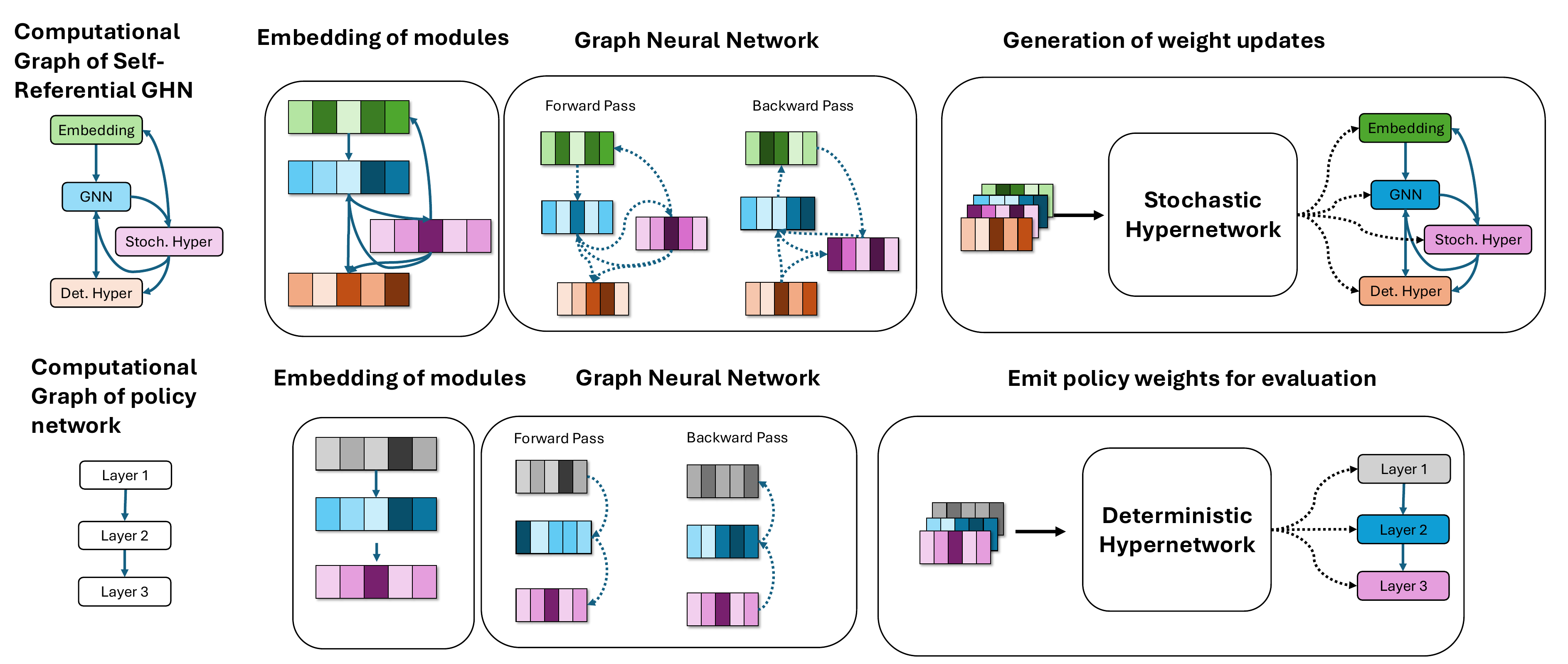}
    \caption{\textbf{Self-Referential Graph HyperNetwork.}  A Graph Hypernetwork produces parameters for another network by considering its computational graph. The GHN learns embeddings for all the node types in the target network and updates these in its graph neural network (GNN) module. These embeddings are then passed on to a hypernetwork module that generates parameters for each node. Self-Referential GHNs have two hypernetwork models: a stochastic hypernetwork that is used to produce updates to copies of the GHN itself, and a deterministic hypernetwork that is used to generate parameters for a target network, in our case, a policy network for reinforcement learning environments. Through this combination, the approach can adapt rapidly to the task at hand. In the figure, for brevity, we depict the GNN, the deterministic, and the stochastic hypernetwork as single nodes, although they each consist of multiple layers that in the experiments are represented as separate nodes in the computational graph. }
    \label{fig:GHN_overview}
\end{figure*}

Yet, most contemporary neuroevolution approaches keep the evolutionary machinery outside the neural substrate being optimized: an algorithm separated from the individuals in the population is responsible for producing mutations to the individuals. In this work, we collapse that boundary by embedding the variation-producing mechanism inside the neural network itself. Building on Graph HyperNetworks (GHNs) \citep{zhang_graph_2020} and prior self-referential systems \citep{ schmidhuber_learning_1992, schmidhuber_steps_1992, chang_neural_2018, randazzo2021recursively}, we introduce \emph{Self-Referential Graph HyperNetworks}: architectures that (i) generate parameter updates to copies of themselves, (ii) inject stochastic, learnable variation into those parameters, and (iii) thereby enact a localized mechanics for increasing and decreasing exploration. See Figure \ref{fig:GHN_overview} for a visual overview of the structure of the self-referential GHN. 

By internalizing evolution, Self-Referential GHNs realize three key advantages:

\begin{enumerate}
\item Derivative-free optimisation of non-differentiable or discrete architectures while remaining fully neural and end-to-end trainable.
\item Rapid adaptation to non-stationary tasks, as demonstrated by our CartPole-Switch and LunarLander-Switch experiments, where populations recover near-optimal performance within a few generations after abrupt controller inversions.
\item Emergent control over mutation magnitude, enabling broad exploration in the early generations and sharp exploitation around high-fitness “discoveries” without any external schedule.
\end{enumerate}

In summary, this paper demonstrates how networks can evolve themselves, not through an external optimiser, but through self-referential mechanisms that reinterpret hypernetwork outputs as heritable, mutable code. As evolvability is a subject to selection in natural evolution \citep{earl2004evolvability}, we believe this type of evolution algorithm is a step toward bridging the gap between artificial evolution and the evolutionary process that has spawned such rich diversity of organisms and intelligences in the real world.



\section{Background}

HyperNetworks \citep{ha_hypernetworks_2016}, a generalisation of HyperNEAT \citep{stanley2009hypercube}, introduced a differentiable small network generating weights for a larger main network, analogous to biological genotype-phenotype mappings. The approach targeted parameter efficiency and relaxed weight sharing across both CNN layers and LSTM timesteps. For LSTMs specifically, this manifested as dynamic weight adaptation (relaxed weight sharing across time), mitigating vanishing gradients common in recurrent models with strict temporal weight sharing. Original applications included sequence modeling in LSTMs and compact CNN kernel generation for image recognition.

Graph HyperNetworks (GHNs) \citep{zhang_graph_2020} extended this by operating directly on the computation graph of the main network, shifting the focus to Neural Architecture Search (NAS) amortisation. GHNs leverage internal graph neural networks \citep{scarselli2008graph, wu2020comprehensive} to process architectural topology and generate parameters for arbitrary, unseen network structures. GHNs serve as fast, surrogate predictors for the evaluation of candidate architectures, without requiring full training.

Following from neural networks parameterising other neural networks, a subsequent line of research has closed this loop via self-referential architectures. \citet{chang_neural_2018} introduced the ``Neural Network Quine,'' a network that can predict its own weights. These networks were tested on auxiliary tasks, such as MNIST classification, and a trade-off between replicative and task performance was found. Notably, Neural Network Quines exhibited infertility after some generations, losing the ability to perform the auxiliary tasks due to imperfect replication. 

As a direct follow-up addressing infertility, \citet{randazzo2021recursively} introduced "recursively fertile" self-replicating neural agents, demonstrating that stability over multiple generations could be achieved by training networks to function as "sinks" in parameter space. This advance meant that the replication process became robust to small perturbations, as slightly perturbed parent parameters would still reliably generate child parameters close to the original, unperturbed state. A key insight was that this stability was enabled by allowing a portion of the network's weights to be perfectly copied (fixed), while others were variably generated by the network itself. This approach was a step towards controlled evolvable diversity, but this variation mechanism was explicitly trained for pre-defined transformations.

More recently, \citet{irie2022modern} proposed Self-Referential Weight Matrices (SRWMs), an extension of earlier concepts of self-referential systems and fast weight programmers \citep{schmidhuber_learning_1992, schmidhuber_steps_1992}. Here, a single weight matrix (WM) acts as both the network's program and the mechanism that modifies this program during runtime. At each step, the SRWM processes an input to produce both the task-specific output as well as the components for its own update. These components are: a modifier-key vector (specifying \textit{what} part of the SRWM to change), an analyser-query vector, used to retrieve a "new value" for the update from the SRWM itself, and a self-generated learning rate. The SRWM then updates itself using a delta rule based on outer products of these self-generated components. Only the initial state of the SRWM is trained via gradient descent. Subsequent modifications are entirely driven by its learned internal dynamics. SRWMs demonstrated the ability to rapidly adapt to changing task environments in few-shot learning and multi-task reinforcement learning.

\citet{shvartzman_self-replicating_2024} introduced Self-Replicating Artificial Neural Networks (SeRANNs). ANNs in this work are encoded with binary genotypes that are translated into Python code defining the network architecture and hyperparameters. Each generated network is then trained via gradient descent to (i) solve a classification task and (ii) copy their own genotype, where copy errors act as mutations leading to evolutionary dynamics. In contrast, the Self-Referential GHNs proposed in this paper change the parameters of copies of themselves, with no gradient updates ever being part of the evolutionary loop.

\section{Self-Referential Evolution of Graph HyperNetworks}


The Self-Referential Graph HyperNetworks are based on the original Graph HyperNetworks (GHNs) developed by \citet{zhang_graph_2020}. GHNs consist of an embedding matrix, a Graph Neural Network (GNN), and an output layer. The GNN takes as input a graph representing the architecture of the neural network for which weights are going to be generated. Each node in this computational architecture is represented by a learned embedding in the embedding matrix. The GNN then updates the representation of each node based on its connectivity in the computational graph. Lastly, each updated node representation is used as input to the output layer, which generates parameters. Note that the output layer always produces the same number of weights, specified by the maximum requirement of all of the nodes. To get the correct number of parameters, $k$, for any smaller node, the first $k$ parameters produced by the output layer are used.

\begin{figure*}[ht]
    \centering
    \includegraphics[width=0.9\linewidth]{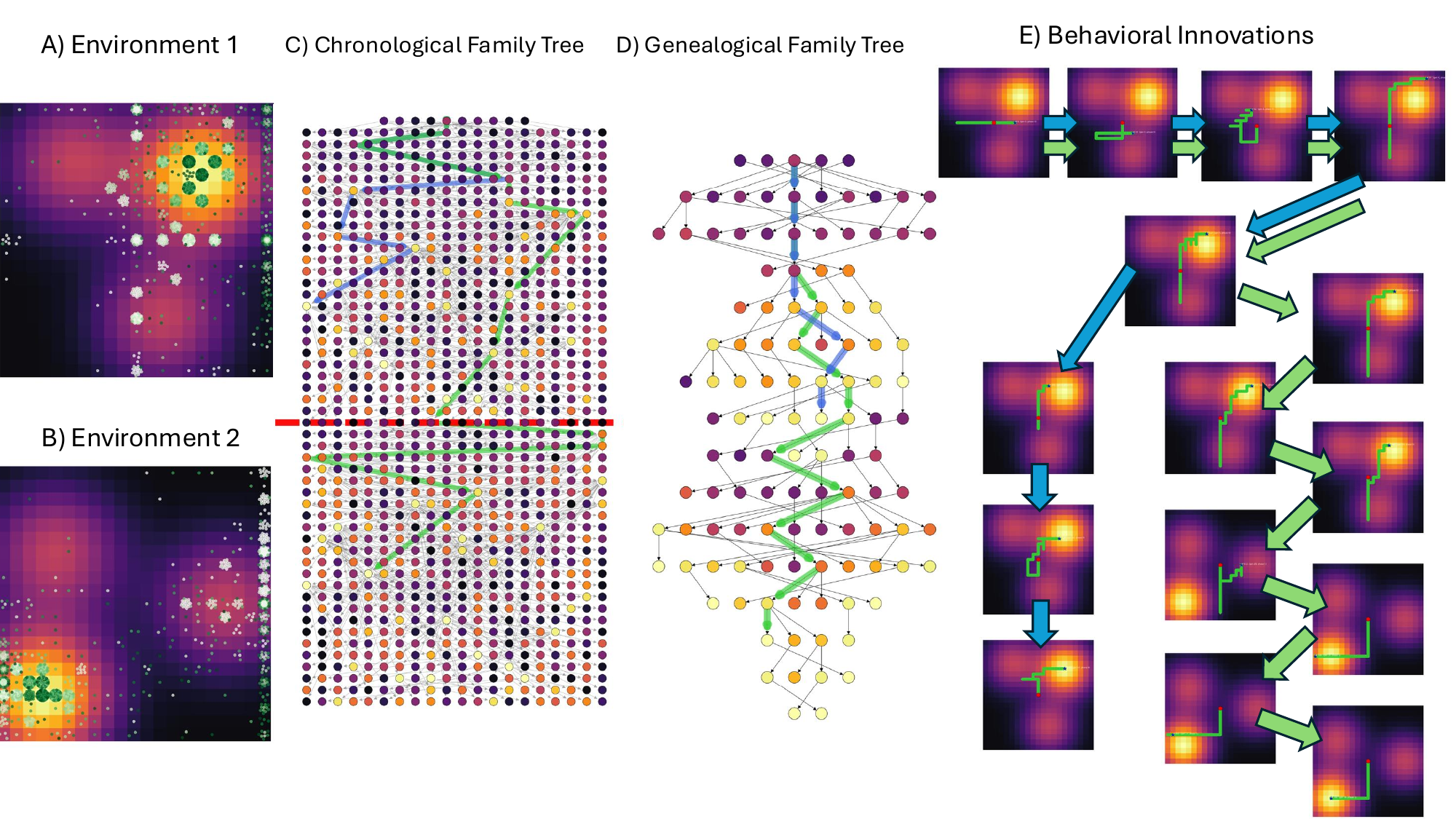}
    \caption{\textbf{2D switching task.} (\textbf{A}) Environment 1: Surface with tiles for navigation (brighter = higher fitness). Points mark where every evaluated policy ended in this phase, with the darker points corresponding to individuals in later generations. Midway through evolution, the landscape switches to Environment 2. (\textbf{B})  Surface with endpoints for all individuals that were evaluated after the shift. (\textbf{C}) Chronological Family Tree: Every dot is an individual; layout follows the order individuals were born. Hue is the individual’s absolute fitness in [0,1]. The blue and green path traces show the lineages of the champions in Environment 0 and 1, respectively. The red divider marks the environment switch. (\textbf{D}) Genealogical Family Tree: Compressed lineage showing only individuals that produced offspring; the horizontal position encodes genealogical generation (parent depth). Colors and champion paths are as in (C). (\textbf{E}) Behavioral Innovations: Ancestor-by-ancestor rollouts for each champion. Each tile shows the actual trajectory (green) taken by that ancestor on the appropriate surfaces, with arrows indicating chronological order. The champions share a common ancestor, but the lineage is split before the environmental shift. These results illustrate the ability of a population of Self-Referential GHNs to adapt to non-stationary tasks.}
    \label{fig:demo}
\end{figure*}

\subsubsection{Stochastic Self-Referential HyperNetwork}
The self-referential GHN follows the same basic structure with a few key differences. The modifications are made to satisfy two basic requirements. The GHN must generate enough parameters to update itself, and these parameters must include some random variation. Both requirements are met through the addition of a stochastic hypernetwork, described in the following. First, to be self-referential, the output layer must be able to generate enough parameters for the GHN itself. To achieve this, we use a fixed random basis as the final output layer. Since this layer never changes, we avoid the circular problem of needing a layer that generates a larger number of parameters than what it itself consists of. 

Second, the original GHNs, once optimized, always generate the exact same parameters given the same neural architecture. Contrary to this, we want the same GHN to generate different sets of candidate weights for the same architecture, such that an evolutionary optimization process can take place. Instead of passing the node representations produced by the GNN directly to the output layer, we use the node representation to generate a parameterization for a distribution, from which a sample is passed to the final output layer. 

Specifically, each node representation from the GNN is sent through a small multi-layer perceptron to produce a set of standard deviations for a multivariate Gaussian distribution with means fixed at zero. With these two modifications, our GHN can generate all the updatable parameters that it itself contains with random variation, which means that we have all that is needed for Self-Referential GHNs that evolve over time.

\subsubsection{Adjustable Rates of Mutation} \label{mutation_rates}

There are some useful properties that we would like to see in a self-referential evolutionary process. Since the parameters of the GHN determine the level of variation in subsequent parameter generation, we would like to facilitate the ability to alter this variation in an opportune manner throughout evolution. Intuitively, if all members of the population achieve roughly the same level of fitness, variation within the population should increase, such that a broader area of the fitness landscape is explored. This condition would be the case at the beginning of the evolutionary phase, when all individuals are likely to perform poorly. On the other hand, when a few individuals in the population perform much better than the rest, we would like the variation within the population to decrease, such that more resources can be concentrated around the more promising solutions.


To let Self-Referential GHNs better control the variability of generated parameters, we predict a node-specific mutation rate that scales the outputs of the stochastic hypernetwork’s final layer. Concretely, each node embedding passes through a linear layer with $M$ sigmoid-activated outputs; the node’s mutation rate is the maximum of these $M$ values.

Choosing the maximum value of multiple possible ones ensures some redundancy in the variation-producing parameters. This mechanism interacts with another simple addition to the stochastic hypernetwork, which is simply to always add a small amount of noise to the final output, \emph{after} the generated parameters have been scaled by the mutation rate.  This means that even if all the $M$ mutation rates are set to zero, slight variations are still possible, including the possibility of increasing the mutation rates in subsequent parameter generation. M is a hyperparameter that controls how easy it is for the Self-Referential GHNs to decrease variation, as well as how easy it is to turn it back on after it has been decreased significantly. In all reported experiments in this paper, we set $M$ equal to five.

\subsubsection{Deterministic Hypernetwork for Fitness Evaluation}
The original purpose of GHNs is to generate parameters for other networks, such that these can solve some task. Self-Referential GHNs follow the same procedure to evaluate the fitness of a GHN in the evolutionary loop. In addition to generating parameters for GHNs in the next generation, each GHN generates parameters for a policy network used to control an agent in an environment. The score achieved in this environment is set as the fitness for the GHN and is used for the selection of individuals in the population.

\subsubsection{Self-Referential Evolution}
We evolve a population of hypernetworks by iteratively letting the top-performing individuals become parents to new offspring. To create an offspring, a parent is first copied exactly. The parent then generates a set of parameters through its stochastic hypernetwork module. This set of parameters is added to the copied parameters of the parent, such that the parent has created a new variation of itself. To generate the new parameters, the parent takes as input its own computational graph. Each node has an embedding, which is also subject to self-referential variation. These embeddings are updated by the graph neural network module of the GHN and are passed on to the stochastic hypernetwork.

For an evolutionary run, a set of Self-Referential GHNs is initialized randomly, independent from each other. Each of these then produces $N$ offspring (in our experiments, $N$ is always equal to two). Next, the fitness of all parents and offspring is estimated. To estimate fitness, each individual generates parameters for a policy network via its deterministic hypernetwork. Here, the input to the GNN is the computational graph of the policy network. The generated parameters are directly assigned to the policy network, and the policy network is deployed to solve a task. An individual's fitness is the score achieved by the policy network.

Figure~\ref{fig:demo} depicts an evolutionary process with the self-referential GHNs. In this toy environment, the agent moves on a 2-D grid where each cell has a fixed “fitness” value. An episode starts at the center; at each step, the policy chooses among up, down, left, right, or wait, and the agent’s fitness is the value of the cell it ends on. The policy network receives eight scalars: the values of the four neighboring cells (up/down/left/right), the value of the current cell, the agent’s x and y coordinates, and a step counter (all normalized). Midway through evolution, the landscape switches to a different one with a new global optimum Figure~\ref{fig:demo}B, so survivors must adapt and navigate to the new target. Each individual is a self-referential Graph HyperNetwork (GHN) that (i) generates the parameters of the policy network and (ii) proposes updates to its own GHN weights to spawn children, enabling an evolutionary loop of replication and adaptation.  This simple task is solvable with a relatively small population size and a few generations. Figure~\ref{fig:demo}C shows the full family tree with every single created individual in chronological order, while Figure~\ref{fig:demo}D depicts a genealogical family tree, where only individuals that were able to produce offspring are included. Here, we can see that the champion individual of the first environment never produced offspring that were successful enough to produce offspring themselves, while a different, parallel lineage eventually results in the champion of the second environment. The population of Self-Referential GHNs demonstrates an ability to quickly build on stepping stones found in the first landscape to find good solutions in the second environment, as seen in Figure~\ref{fig:demo}E.

\section{Experiments}

\begin{figure}[htbp!]
    \centering
    \includegraphics[width=0.8\linewidth]{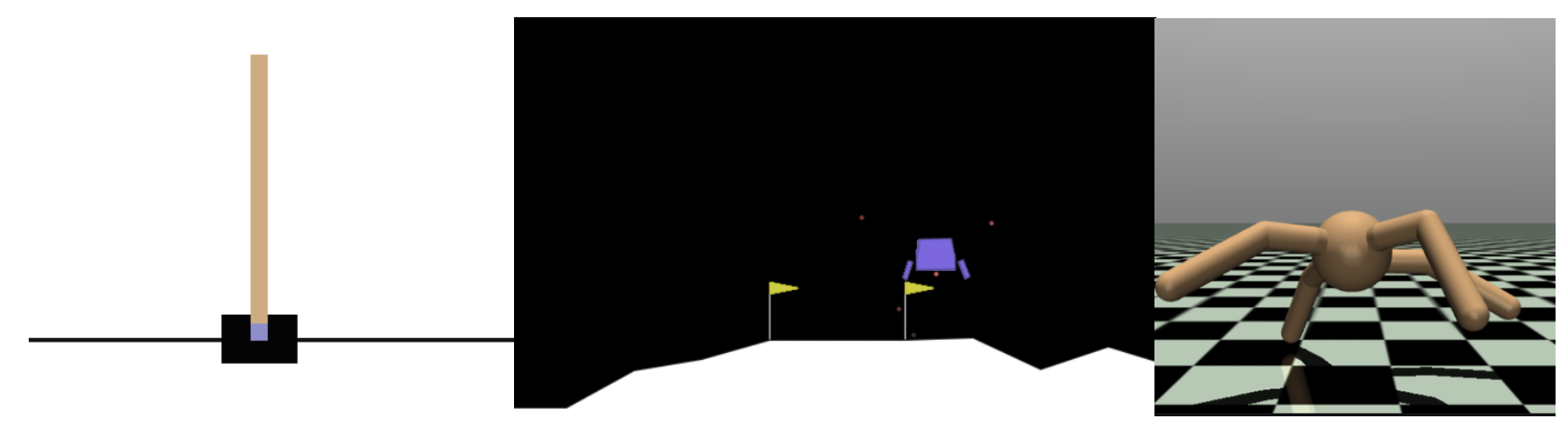}
    \caption{\textbf{Environments} Right: CartPole used for the CartPole-Switch task. Middle: LunarLander used for the LunarLander-Switch task. Right: Ant-v5.}
    \label{fig:envs}
\end{figure}


Before describing the environments in which we test our proposed approach, we describe some hyperparameters of the Self-Referential GHNs that are common across all our experiments. First, for numerical stability, when an offspring is created, we clip its parameters to have values between -20 and 20. Similarly, we clip the magnitude of the standard deviations to be between zero and two. The outputs of the stochastic hypernetwork are clipped to be between -0.1 and 0.1 before being multiplied with the produced mutation rate. The standard deviation of the constant noise that is added to the outputs of the stochastic hypernetwork is set to 0.001. 
Further, all experiments used a node embedding size of 32 for representing the nodes for which the parameters are produced.

These hyperparameters were not found through an extensive search, but were simply found to be appropriate in preliminary experiments.

\subsubsection{CartPole-Switch}
This environment builds on the classic control environment, CartPole \citep{sutton1995generalization} (Figure~\ref{fig:envs}, left). Here, an agent has to move a cart left and right to prevent a pole from falling. Agents are evaluated on many time steps they manage to balance the pole, a perfect score being 500 time steps. The CartPole-Switch task begins exactly as the original environment. However, at generation 600, the outputs of the agent are interpreted oppositely; an action that before resulted in moving the cart to the left now moves the cart to the right, and vice versa. At generation 1,000, the output interpretation is switched back to the original. The full evolution run is 1,500 generations.

We choose output switching as the change in the environment that the evolution algorithm needs to recover from, as this ensures that the behavior of the agent truly needs to change. The inputs to the agent remain the same, but the agent must respond to these differently.

The CartPole-Switch task provides a light-weight setup in which we can compare several different evolutionary algorithms in their ability to recover from their populations' performance after a change occurs in the environment.

We compare our method with classic evolution strategies: OpenES \citep{salimans2017} and CMA-ES \citep{hansen_adapting_1996}, as well as a simple genetic algorithm with point-crossover and a decaying mutation rate. We also compare to methods employing mutation rate adaptation. Group Elite Selection of Mutation Rates (GESMR) co-evolves a population of solutions and a population of mutation rates that are then associated ans separetely selected \citep{kumar_effective_2022}. Self-Adaptation Mutation Rates (SAMR), on the other hand, evolves a population of solutions where each solution corresponds to both parameters and their associated mutation rates \citep{clune_natural_2008}. We use implementations provided in the Evosax package \citep{lange_evosax_2022} for both GESMR and SAMR.

Common to all approaches is that they evolve a policy network, which is a small MLP with a single hidden layer of 32 neurons that are activated by the hyperbolic tangent function. Further, all population sizes are set to 30, and all genetic algorithms have an elite population size of ten.

A hyperparameter specific to the Self-Referential GHNs is that they, in this task, produce five mutation rates to choose from (see Section~\ref{mutation_rates} "\nameref{mutation_rates}").

\subsubsection{LunarLander-Switch}
To further test our Self-Referential GHNs, we modify the LunarLander-v3 environment \citep{towers2024gymnasiumstandardinterfacereinforcement} (Figure~\ref{fig:envs}, middle) in a similar manner as we did the CartPole environment above. The first 600 generations proceed as in the standard LunarLander environment, where the agent is tasked with landing a small rocket securely on the ground while spending as little fuel as possible. At each time step, the agent must choose between four discrete actions: 1) do nothing, 2) fire the left orientation engine, 3) fire the main engine, or 4) fire the right orientation engine. The agent scores extra points by landing in a designated goal area. An episode is considered successful with a score of above 200 and with the maximum possible score being slightly above 300.

At generation 600, we reverse the ordering of the agent's outputs, to become 1) fire the right orientation engine, 2) fire the main engine, 3) fire the left orientation engine, and 4) do nothing. At generation 1,000, the ordering is switched back to the original ordering.

The LunarLander environment is significantly harder than the CartPole. It thus serves as a further verification of the abilities of the Self-Referential GHNs to evolve and adapt.

In this environment, the policy network that GHNs produce parameters for are MLPs similar to the ones in the CartPole-Switch task, but with two hidden layers of 32 neurons instead of one. The population size is increased to 300 with an elite population size of 100. Since LunarLander is a noisy environment with large variation across episodes, the fitness of an individual is set to be the average score of two independent episodes.

\subsubsection{Ant-v5}
Beyond testing for changes in the fitness landscape, we also test the Self-Referential GHN approach in its ability to optimize a policy for a more difficult locomotion problem with continuous outputs: the Ant-v5 \citep{towers2024gymnasiumstandardinterfacereinforcement} (Figure~\ref{fig:envs}, right). In this environment, the agent must control a four-legged robot, learning to walk from scratch and steering it in the correct direction. We modified this environment to only use the first 27 dimensions of the observation space out of 107 available dimensions. These inputs provide proprioceptive information to the agent and are sufficient for learning locomotion.

For this environment, the policy network is an MLP with three hidden layers of 32 neurons. All neurons are activated by the hyperbolic tangent function. The population size is 150, with an elite population size of 50, meaning that each surviving individual produces two offspring.

In this task, only a single mutation rate is produced in the stochastic hypernetwork. Further, instead of clipping the outputs of the stochastic hypernetwork to be between -0.1 and 0.1, they are clipped to be between -1 and 1.

Here, the GHNs are used to produce parameters for two different networks: the policy network, which has a fixed architecture throughout all evolution runs, and GHNs themselves, which also have fixed architectures.

\section{Results}

\subsubsection{CartPole-Switch}

\begin{figure*}[ht]
    \centering
    \includegraphics[width=0.9\linewidth]{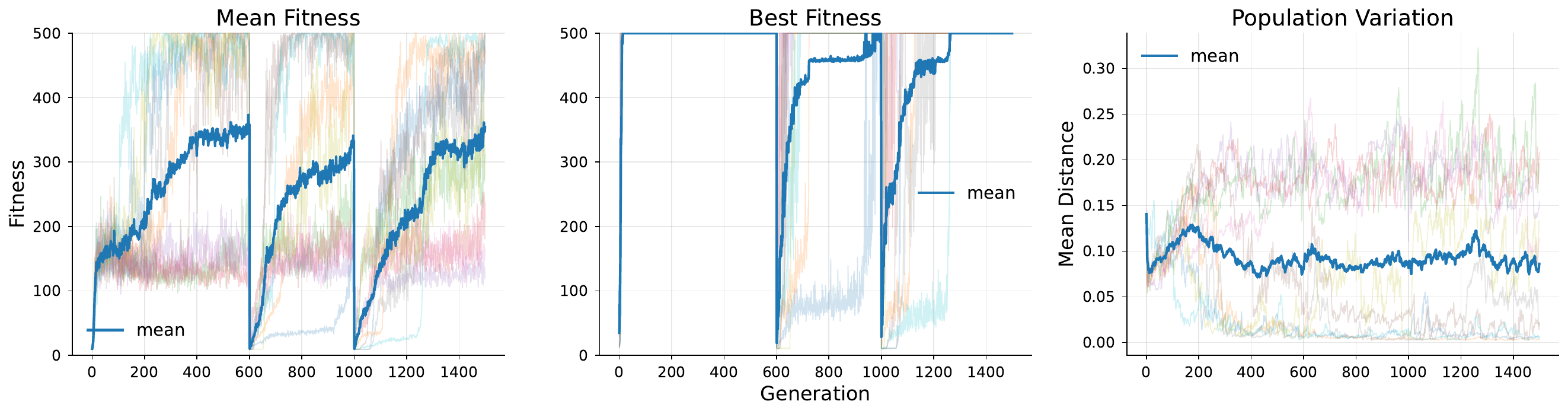}
    \caption{\textbf{CartPole-Switch:} Self-Referential GHNs performance in the CartPole-Switch environment. The curves for ten different evolution runs with different seeds are shown, with the mean curve overlaid (opaque blue lines). In all ten runs, the Self-Referential GHNs recover the performance after both switches, such that the best-performing individuals in the population score the highest possible fitness. }
    \label{fig:cart_ghn}
\end{figure*}

Results are shown for ten different runs with different random seeds for each of the tested algorithms. Training curves exhibiting the mean population scores, the scores of the top-performing within the population, as well as the averaged distance between each pair of individuals within the population of the Self-Referential GHNs can be found in Figure \ref{fig:cart_ghn}. 

There is a large variation across runs in how much individuals differ from each other as measured by the average pair-wise Euclidean distance in the population. However, the distance curves are always characterized by having peaks early in evolution and soon after each switch. 

In all runs, we see a sharp decline in performance immediately following a switch. Individuals with perfect scores are eventually found within each population. In the majority of runs, this happens after only a few generations.

\begin{figure*}[ht]
    \centering
    \includegraphics[width=0.9\linewidth]{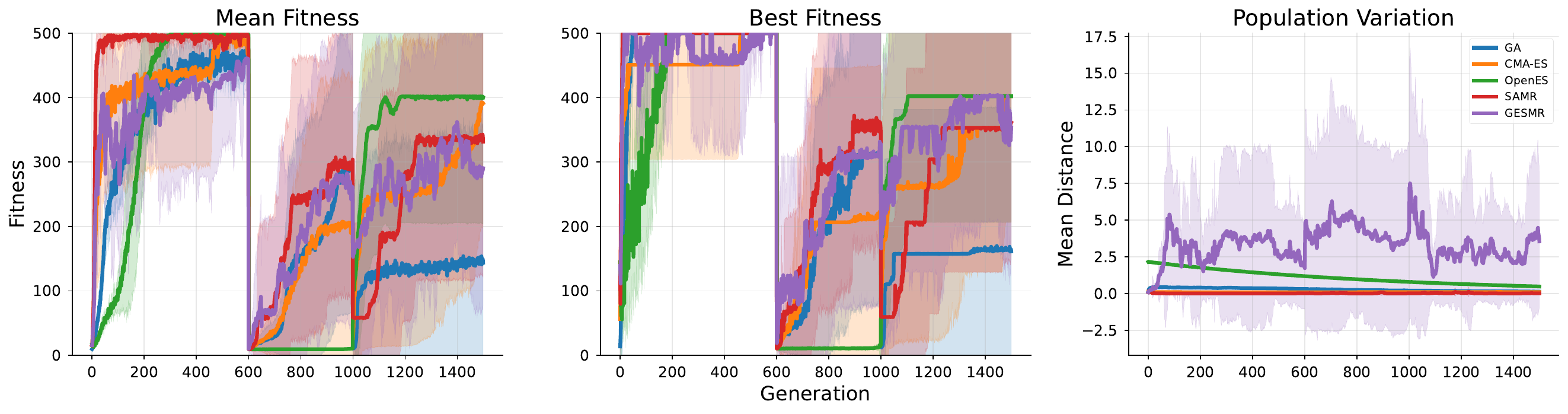}
    \caption{\textbf{CartPole-Switch:} Performance of evolution algorithms used for comparison in the CartPole-Switch environment. None of the algorithms can consistently recover their performance after both switches. The GESMR algorithm produces populations with large amounts of variation compared to all other algorithms. However, even though the GESMR algorithm can introduce large amounts of variation closely after the environmental changes, this algorithm also fails at recovering performance.}
    \label{fig:cart_controls}
\end{figure*}

This is in contrast to the performance of the populations of the other evolution algorithms that we tested in the same environment. Means and standard deviations of the training curves are shown in Figure \ref{fig:cart_controls}. While all algorithms find high-performing individuals within the first 600 generations, full recovery after outputs have been switched is rare, not to mention full recovery after \emph{both} switches.

The average pair-wise distance of GESMR dwarfs the distances measured for any of the other algorithms. This happened even though the parameters of the policies evolved by the GESMR algorithm were clipped to be between -20 and 20, the same as for the GHNs. However, even with these large variations, the GESMR algorithm failed to consistently recover performance after both switches.

\begin{figure}[!ht]
    \centering
    
    \includegraphics[width=1.\linewidth]
    {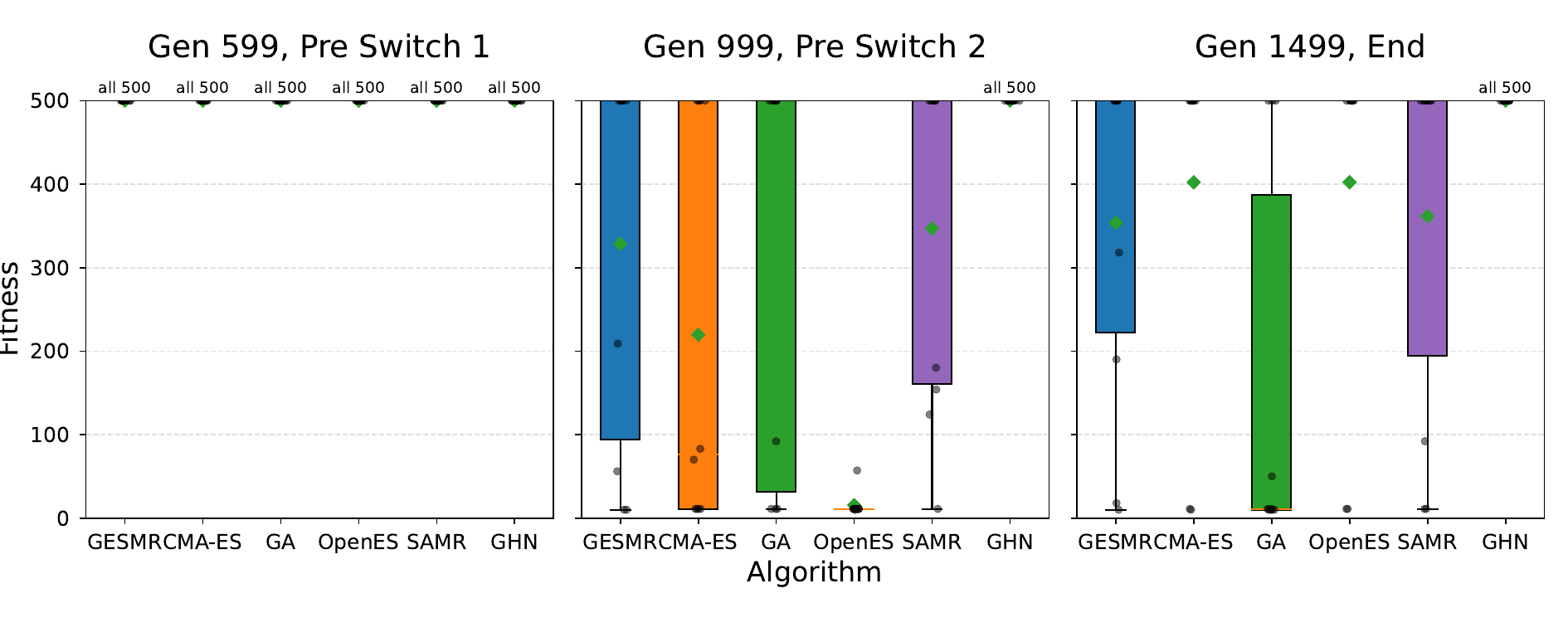}
    \caption{\textbf{CartPole-Switch:} Box plots of best performances of each algorithm before and after switches. All algorithms have their top-performing individuals achieve the highest possible score before the switch. However, only the Self-Referential GHNs consistently have top-performers that achieve this score after both switches. Note that while CMA-ES and OpenES seem to have nearly perfect performance by the end of the evolution after the second switch, that does not equate to having successful runs. Indeed, OpenES never recovers from the first switch, meaning that good performance after the second switch can not be seen as a "recovery" or performance.}
    \label{fig:cart_box}
\end{figure}

The differences in the performances before and after switches are emphasized in Figure \ref{fig:cart_box}, where box plots show the performance across the 10 runs of each algorithm right before a switch and at the end of the evolution run. Only the Self-Referential GHNs achieve perfect scores in each of the points of measurement.

The OpenES algorithm seems to have perfect scores after the second switch. However, even its best individuals have abysmal performances right before the second switch, indicating that it never managed to steer its population away from the solutions that it found after the first 600 generations in the original task setting.

\subsubsection{LunarLander-Switch}

\begin{figure}[!ht]
    \centering

    \includegraphics[width=1.\linewidth]{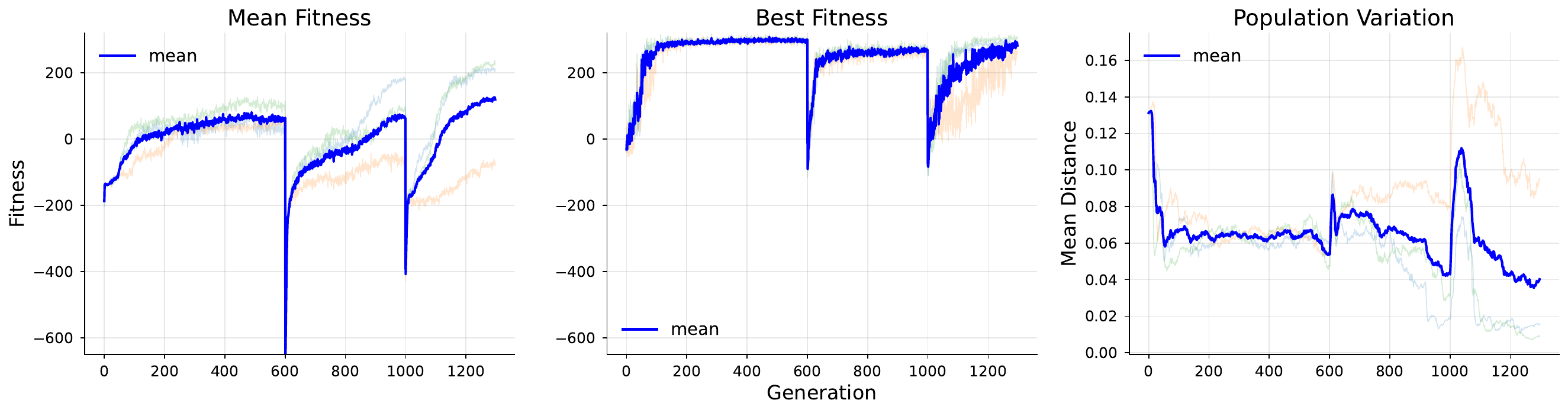}
    \caption{\textbf{LunarLander-Switch: } Self-Referential GHNs performance in the LunarLander-Switch environment. The top-performing individuals quickly achieve high scores after switches occur. Peaks in population variation seem to follow the switching events closely.}
    \label{fig:lunar}
\end{figure}

In the LunarLander-Switch environment, we see a similar tendency for the Self-Referential GHNs as in the CartPole-Switch environment (Figure~\ref{fig:lunar}). Across three runs with different random seeds, the GHNs quickly recover from both changes to the output ordering. The changes in population variation is even more pronounced in this environment with large peaks close right after the switches.

\subsubsection{Ant-v5}
\begin{figure}[ht]
    \centering
    
    \includegraphics[width=1.\linewidth]
    {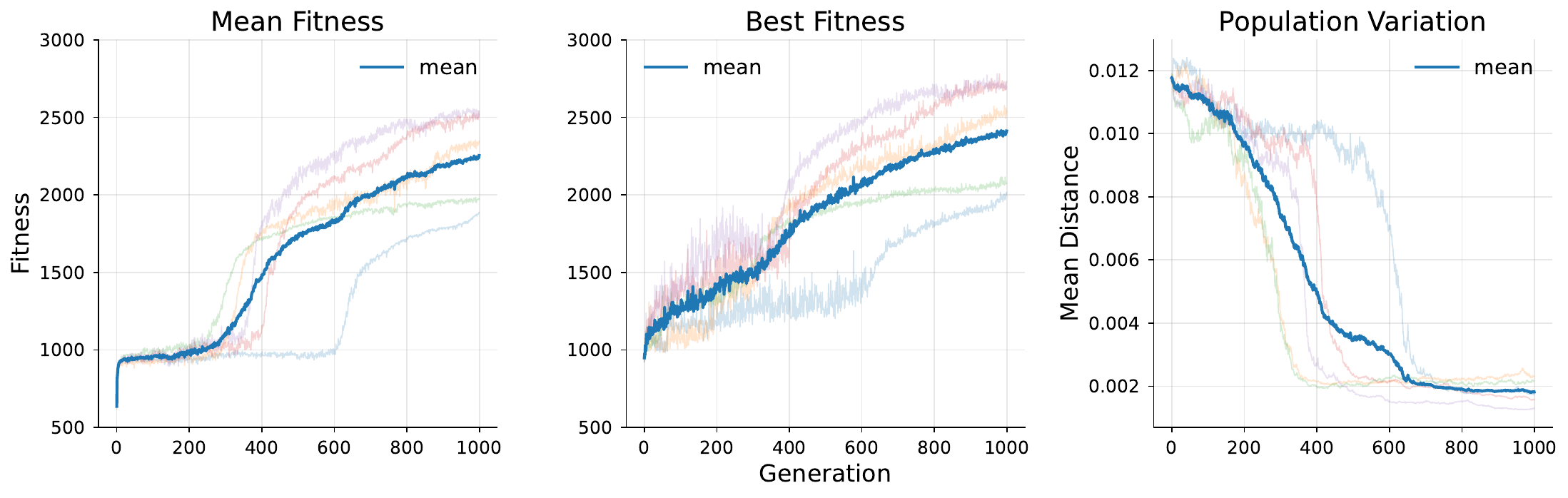}
    \caption{\textbf{Self-Referential GHNs performance in the Ant-v5 environment.} While the runs do not achieve the maximum score within the time limit of 1,000 generations, we see clear improvements over time. Increases in the mean fitness of the population are associated with a drop in the population variation.}
    \label{fig:ant}
\end{figure}

 In this task, the fitness landscape remains constant, and the performance in each run improves monotonically (Figure~\ref{fig:ant}). The runs vary in how soon in the evolution they get past the local optimum point with a score of around 1,000, where the robot stands perfectly still, and simply avoids penalties from falling. In each case, a break-through in performance away from this local optimum is followed by a sharp decrease in the variation within the population.

Wihtin the budget of 1,000 generations allotted to the evolution process, none of the runs achieved the highest possible score, which is around 4,000 points in the Ant-v5 environment. However, with consistent scores above 2,000, the Self-Referential GHNs have produced agents capable of locomotion.
Further, the training curves have not yet converged within, and it is thus possible that a larger budget could result in even better scores.

\section{Discussion and Future Perspectives}

This work has shown that Self-Referential Graph HyperNetworks can “evolve themselves”.
By embedding a stochastic hypernetwork inside the agent, each Self-Referential GHN produces child weight sets that contain learnable, state-dependent noise, while a deterministic hypernetwork generates the policy weights whose performance is used as fitness.
Across three benchmarks, the approach consistently (i) recovers almost instantly after abrupt controller inversions in CartPole-Switch and LunarLander-Switch, achieving perfect or near-perfect scores within a handful of generations, and (ii) evolves promising locomotion on Ant-v5.
Throughout, population-level diversity rises after environmental changes and decreases once high-fitness niches are found, indicating emergent control of mutation magnitude. 

Internalizing variation offers two advantages.
First, because mutation rates are themselves heritable parameters, selection can directly favour individuals whose exploratory behaviour matches the current ruggedness of the fitness landscape, echoing the “evolvability as a selectable trait” hypothesis from biology \citep{earl2004evolvability}.
Second, separating the where (deterministic hypernetwork) from the how much (stochastic hypernetwork) encourages a division of labor: once a promising policy is discovered, the mutation-rate subnet shuts exploration down, funneling search effort into fine-tuning nearby.
The peaks in pair-wise Euclidean distance immediately after each switch, especially apparent in Figure \ref{fig:lunar},  and the sudden contraction that follows after breakthroughs in performance Figure \ref{fig:ant} demonstrate this self-regulated exploration–exploitation cycle.

Placing the mechanism of mutation-generation as learnable modules within individuals comes with some downsides. Namely, generating parameter updates is a much more computationally costly process than simply drawing the mutations from a distribution. Further, the size of the deterministic hypernetwork is determined by the maximum parameter requirement of any computational node in the target policy network that we want to generate parameters for. In turn, the size of the stochastic network is determined by the maximum parameter requirement of any computational node within the rest of the GHN, most often the output layer of the deterministic hypernetwork. This means the size of the Self-Referential GHN grows quickly with the size of the target networks, making the process of parameter generation even slower.
Future work should address this, for example, by replacing the updatable output layer of the deterministic hypernetwork with a random basis, just as we did with the output layer of the stochastic hypernetwork in this paper.

Another avenue of future research will be self-referential evolution with multiple target networks, instead of just a single one as was done in the experiments above. The Self-Referential GHNs maintain the potential of the original GHNs \citep{zhang_graph_2020} for generating parameters for neural networks with differing architectures, even ones not seen during the training phase.
Exploring this also opens the door for neural architecture search \citep{liu2021survey} with mutations being applied to the neural architectures of the target networks as well as their parameters. In the extreme case, the Self-Referential GHNs could even mutate their own neural network architectures, creating even larger potentials for evolving high rates of diversity within the population. A possible direction is to combine the GHNs with a Neural Developmental Program \cite{najarro2023towards, plantec2024evolving} such that development and evolution can interact.   


By demonstrating that purely neural systems can house replication and variation internally, this work takes a step toward being a closer analogy to natural evolution. We hope that this work will inspire the development of further research into novel evolution algorithms and neural networks that can improve themselves.

\section{Acknowledgments}
Funded by the European Union (ERC, GROW-AI, 101045094). Views and opinions expressed are however those of the authors only and do not necessarily reflect those of the European Union or the European Research Council.

\footnotesize
\bibliographystyle{apalike}
\bibliography{references} 

\end{document}